\documentclass[aps, prd, superscriptaddress, twocolumn, nofootinbib, floatfix]{revtex4-2}

\usepackage{amsmath}
\usepackage{amssymb}
\usepackage{graphicx}
\usepackage{xcolor}

\usepackage[colorlinks,linkcolor=blue,citecolor=blue,urlcolor=blue]{hyperref}

\newcommand{\sNN}{\sqrt{s_{NN}}}
\newcommand{\dd}{\mathrm{d}}

\begin{document}

\title{Effects of Parton Pair Correlations on Direct-Photon Yields and Elliptic Flow in Heavy-Ion Collisions}

\author{Xingjian Lu}
\email[]{lus21@tsinghua.org.cn}
\affiliation{School of Physics and Electronic Information, Yantai University, Yantai 264005, China}
\affiliation{Department of Physics, Tsinghua University, Beijing 100084, China}

\author{Shuzhe Shi}
\email[]{shuzhe-shi@tsinghua.edu.cn}
\affiliation{Department of Physics, Tsinghua University, Beijing 100084, China}

\begin{abstract}
Direct photons escape heavy-ion collisions with little rescattering, carrying information from the full emission history.
Yet describing the measured direct-photon yield and the sizable elliptic flow $v_2$ together remains a long-standing challenge.
Here we explore how correlations between emitting partons may affect both observables.
The local $2\to2$ photon rate depends on the joint distribution of incoming partons, so a connected correlation between them can change photon emission without changing the one-particle distribution of medium particles.
Once folded over the evolving medium in Au+Au and Pb+Pb collisions, the correlation raises the photon yield toward the fitted PHENIX and ALICE data and increases $v_2$ mainly at intermediate $p_T$.
The fitted temperature dependence accounts for most of the net flow increase in this region, suggesting that the correlation changes the relative contributions between hotter and cooler regions.
\end{abstract}

\maketitle

\section{Introduction}
\label{sec:introduction}

Photons are produced at every stage of a heavy-ion collision and then leave the medium without further interaction~\cite{Feinberg:1976ua, Shuryak:1978ij, Kajantie:1981wg, McLerran:1984ay, Kajantie:1986dh, Tserruya:2009zt}. Their spectrum therefore carries an integrated record of the temperature and of the collective flow of the matter that emitted them. Two observables carry this record. The azimuthally averaged yield measures how much radiation the evolving medium produces, and the elliptic flow coefficient $v_2$ measures how strongly this radiation follows the anisotropic expansion. See~\cite{Gale:2025ome, Jackson:2025mtk, Du:2025ith} for recent reviews and~\cite{PHENIX:2008uif, PHENIX:2011oxq, Liu:2012ax, Rapp:2014hha, ALICE:2018dti, Dasgupta:2020orj, Sun:2023rhh, Stankus:2005eq, Liu:2007tw, Vitev:2008vk, David:2019wpt, Stankus:2005eq, Liu:2007tw, Vitev:2008vk, David:2019wpt, Owens:1986mp, Churchill:2020uvk, Garcia-Montero:2023lrd, Garcia-Montero:2024lbl, Paquet:2015lta, Turbide:2003si, Fries:2002kt, Turbide:2005fk, Qin:2009bk, Wu:2024vyc, Gao:2026vxs, Du:2025dot, Churchill:2023zkk} for related studies. The PHENIX and STAR collaborations have measured direct photons in Au+Au collisions at $\sNN=200~\mathrm{GeV}$~\cite{PHENIX:2011oxq, PHENIX:2014nkk, PHENIX:2015igl, STAR:2016use, PHENIX:2022rsx} performed in the Relativistic Heavy-Ion Collider (RHIC), and the ALICE collaboration has measured them in $\sNN=2.76~\mathrm{TeV}$ Pb+Pb collisions~\cite{ALICE:2015xmh, ALICE:2018dti} at Large Hadron Collider (LHC). In both systems the measured photon $v_2$ is sizable; at RHIC it is of the same order as the hadron $v_2$~\cite{PHENIX:2011oxq}, although a large part of the thermal radiation is expected from early, hot, and still weakly anisotropic stages. Describing a large yield and a large $v_2$ with one model is the long-standing direct-photon problem~\cite{David:2019wpt, Paquet:2015lta}.

The central difficulty is to describe the direct-photon yield and $v_2$ with the same calculation.
The thermal photon signal mixes emission from a hot, weakly anisotropic early medium with emission from a cooler later medium in which stronger collective flow has developed.
Increasing the early contribution can raise the yield but dilute $v_2$, while giving more weight to late emission can raise $v_2$ but also change the yield and its momentum dependence.
This interplay makes it difficult to describe both observables with one set of photon sources.
A recent multistage calculation that combines prompt, pre-equilibrium, and hydrodynamic thermal photons~\cite{Gale:2021emg} illustrates the difficulty in the data considered here.
In the low-$p_T$ region, the calculation comes closest to the STAR yield but undershoots the PHENIX and ALICE yields.
Its $v_2$, however, lies below the measured central values in nearly all fitted bins.
Therefore, a viable mechanism must improve $v_2$ and reduce the yield deficit in the bins where the baseline is insufficient, while preserving the agreement in the bins where the yield is already well described.

Most photon-rate calculations assume that the colliding partons are uncorrelated, so their two-particle distribution is a product of one-particle distributions~\cite{Arnold:2001ms,Paquet:2015lta}.
In a dense, strongly interacting medium, this assumption may fail~\cite{Lu:2025yry}.
In the $2\to2$ annihilation and Compton processes, a connected two-particle correlation enters the photon emission rate linearly and can produce an oscillatory correction to the local photon spectrum~\cite{Lu:2026nmp}.
This correction can enhance emission at some photon momenta and suppress it at others.
When folded over a flowing medium, this momentum-dependent correction can affect the yield and $v_2$ differently.
In Ref.~\cite{Lu:2026nmp}, we established this local spectral effect but did not fold it over a realistic space-time evolution.
It therefore left open whether the effect can describe measured yields and $v_2$, and which parts of the correlation the data constrain.

Here we ask whether a local pair-correlation response can describe the direct-photon yield and $v_2$ in Au+Au and Pb+Pb collisions when the other photon sources and hydrodynamic inputs are held fixed.
We study the $0-20\%$ and $20-40\%$ centrality classes in both systems.
We expand the correlation in a truncated momentum-space basis with two spatial features, giving $146$ candidate coefficients, of which four are fitted, and reconstruct the coefficients separately for each class.
We assess which combinations of coefficients the data constrain.
We also test whether the total local photon emission rate is nonnegative in all cells above $T_{\rm sw}$ at sampled photon energies,  whether the total yield is positive in every experimental bin, whether the two-particle distribution is nonnegative in a weighted domain, and how these conditions affect the fit.

The result of the paper is organized as follows. Section~\ref{sec:method} presents the model and method, Sec.~\ref{sec:results} the results, and Sec.~\ref{sec:conclusion} the conclusions.

\section{Model and method}
\label{sec:method}

\subsection{Local photon production rate}
\label{sec:method_rate}
For completeness, we briefly review the correlation induced photon production calculation developed in Ref.~\cite{Lu:2026nmp}.
For a tree-level process $a(p_a)+b(p_b)\to c(p_c)+\gamma(k)$ in a medium cell at space-time point $X$, the invariant photon emission rate is
\begin{align}
 E_k\frac{\dd R}{\dd^3k}(X)
 ={}&\sum_{ab\to c\gamma}\frac{1}{2(2\pi)^3}
 \int \dd\Pi_a\,\dd\Pi_b\,\dd\Pi_c
 \nonumber\\
 &\times(2\pi)^4\delta^{(4)}(p_a+p_b-p_c-k)
 \nonumber\\
 &\times\overline{|\mathcal M|^2}_{ab\to c\gamma}\,
 \mathcal S_{abc}(X;p_a,p_b,p_c),
 \label{eq:rate}
\end{align}
where $\dd\Pi_i=\dd^3p_i/[(2\pi)^3 2E_i]$ and $\overline{|\mathcal M|^2}$ contains the squared matrix element with the sums and averages over internal quantum numbers. For massless $u$, $d$, and $s$ quarks the channel sum contains $q\bar q\to g\gamma$ and $q(\bar q)g\to q(\bar q)\gamma$. The medium enters only through the statistical factor $\mathcal S_{abc}$.

To separate genuine correlations from single-particle effects, we write the two-particle distribution of species $i$ and $j$ as
\begin{equation}
 \mathcal F_{ij}(X;p_i,p_j)=f_i(X;p_i)f_j(X;p_j)+g_{ij}(X;p_i,p_j),
 \label{eq:pair_decomposition}
\end{equation}
and define the connected part $g_{ij}$ by
\begin{gather}
 \int \dd^3p_i\,g_{ij}=0,\qquad \int \dd^3p_j\,g_{ij}=0,
 \nonumber\\
 g_{ij}(X;p_i,p_j)=g_{ji}(X;p_j,p_i).
 \label{eq:zero_marginals}
\end{gather}
The two zero-marginal conditions ensure that $g_{ij}$ does not change the one-particle distributions. Without genuine three-particle correlations,
\begin{equation}
 \mathcal S_{abc}=f_af_b(1+\sigma_cf_c)
 +\big[g_{ab}(1+\sigma_cf_c)+\sigma_cf_bg_{ac}+\sigma_cf_ag_{bc}\big],
 \label{eq:statistical_factor}
\end{equation}
with $\sigma_c=+1$ ($-1$) for a bosonic (fermionic) final-state parton. We use the classical-statistics, leading-logarithmic limit of Ref.~\cite{Lu:2026nmp}, in which $\mathcal S_{abc}\to f_af_b+g_{ab}$. The correlation correction to Eq.~\eqref{eq:rate} is then linear in $g_{ab}$. We assume the same momentum-space correlation profile for all parton species.

In the local rest frame of a rotationally invariant cell, $g_{ab}$ can depend only on the two energies scaled by the temperature, $\xi_i=p_i\!\cdot\!u(X)/T(X)$, and on the relative angle, $z_{ab}=\hat{\boldsymbol p}_a\cdot\hat{\boldsymbol p}_b$. Rotational invariance removes any preferred absolute direction, but it leaves correlations in $z_{ab}$. We expand
\begin{align}
\begin{split}
 &g_{ab}(X;p_a,p_b)
 \\=\;&\sum_{n,m,\ell}\Psi_{nm\ell}(X)\,f^{\rm ref}_a f^{\rm ref}_b
 L^{(2)}_n(\xi_a)L^{(2)}_m(\xi_b)P_\ell(z_{ab}),
\end{split}
 \label{eq:basis}
\end{align}
with the Boltzmann reference distribution $f^{\rm ref}_i=e^{-\xi_i}$, generalized Laguerre polynomials $L^{(2)}_n$, and Legendre polynomials $P_\ell$.

A mode in Eq.~\eqref{eq:basis} is a connected correlation only if it satisfies Eq.~\eqref{eq:zero_marginals}. Integrating over $p_b$ gives
\begin{align}
\begin{split}
&
    \int\dd^3p_b\, f^{\rm ref}_bL^{(2)}_m(\xi_b)P_\ell(z_{ab})
\\=\,&
    4\pi T^3\delta_{\ell0} \int_0^\infty \dd\xi\,\xi^2 e^{-\xi}L^{(2)}_m(\xi)
\\=\;&
8\pi T^3\delta_{\ell0}\delta_{m0}.
\end{split}
 \label{eq:marginal}
\end{align}
The radial integral equals $2\delta_{m0}$ by the orthogonality of $L_m^{(2)}$ with the weight $\xi^2e^{-\xi}$.
Every product mode with $\ell\ge1$ therefore satisfies both zero-marginal conditions.
For $\ell=0$, a single product mode satisfies both conditions when $n,m\ge1$.

\begin{figure*}
 \centering
 \includegraphics[width=6.0in]{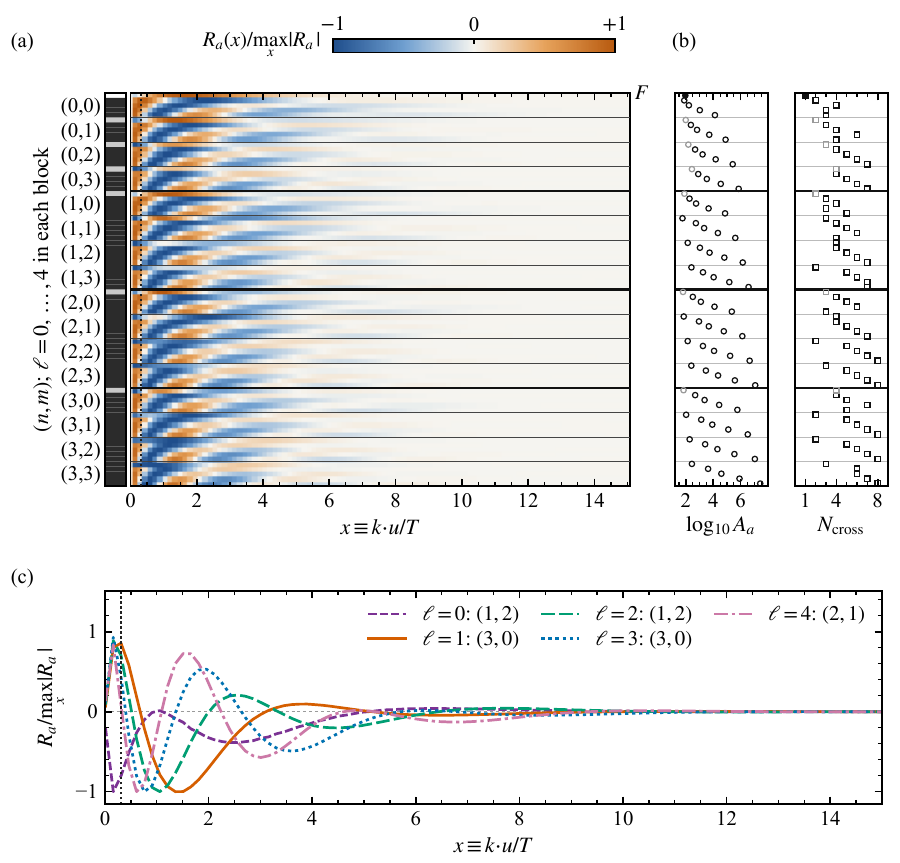}
 \caption{
 Local response library for the pair-correlation basis of Eq.~\eqref{eq:basis}. (a) Stored responses $R_a(x)\equiv y_{nm\ell}(x)$ in the order $(n,m,\ell)$, with $\ell=0,\ldots,4$ inside each $(n,m)$ block, as functions of $x\equiv k\!\cdot\!u/T$. Each row is divided by its own $\max_x|R_a|$; the color shows sign. The strip on the left marks in dark gray the $73$ connected modes, the $64$ modes with $\ell\ge1$ and the nine $\ell=0$ modes with $n,m\ge1$, which satisfy Eq.~\eqref{eq:zero_marginals} and form the candidate pool of the fit; the six $\ell=0$ modes with $nm=0$ have nonzero marginals and are excluded (light gray). The top row is the factorized reference $F$. The dotted line at $x=0.31$ is the lower edge of the fitted response range. (b) Logarithm of the stored amplitude $A_a=\max_x|R_a|$ and number of sampled sign changes (threshold $10^{-6}A_a$). Open black, open gray, and filled symbols denote connected modes, excluded modes, and $F$, respectively. (c) For each $\ell=0,\ldots,4$, the connected mode whose normalized shape has the smallest mean squared distance to the other connected modes with the same $\ell$; the labels give $(n,m)$.
 }
 \label{fig:atlas}
\end{figure*}

The amplitude of each mode may vary across the medium. We write
\begin{equation}
 \Psi_{nm\ell}(X)=\sum_A\psi_{A,nm\ell}\,q_A(X),
 \label{eq:features}
\end{equation}
where $\psi_{A,nm\ell}$ are global coefficients, $q_A(X)$ are fixed features of the hydrodynamic fields, and $A$ labels the feature.
We use two features: a constant that is independent of the hydrodynamic property, $q_{\rm C}=1$, and the logarithm of the local temperature which captures the difference in correlation caused by the density related collision rates, $q_T=\ln[T(X)/0.2~\mathrm{GeV}]$.
The reference value $0.2~\mathrm{GeV}$ defines the zero of $q_T$. Its exact value is not very important, because changing it only adds a constant that can be absorbed into the corresponding $q_{\rm C}$ coefficient.
Both features are rotational scalars with the same definition in every hydrodynamic profile.
With the $73$ connected modes of the library (Fig.~\ref{fig:atlas}) and two features, the candidate set has $73\times2=146$ coefficients; for each class we fit four of them (Appendix~\ref{app:inverse}).

Because the correction is linear in $g_{ab}$, each mode produces a local spectral response $\mathcal R_{nm\ell}(x,T)$, the change of the invariant rate per unit mode amplitude, where $x\equiv k\!\cdot\!u/T$ is the photon energy in the local rest frame in units of $T$. The correlation contribution to the local rate is
\begin{equation}
 \Delta\Gamma^{\rm pair}_\gamma(X;k)=\sum_{A,n,m,\ell}\psi_{A,nm\ell}\,q_A(X)\,\mathcal R_{nm\ell}\!\left(\frac{k\!\cdot\!u}{T},T\right).
 \label{eq:local_increment}
\end{equation}
Instead of an incomplete leading-order rate in the demonstration study~\cite{Lu:2026nmp}, we use the phenomenological additive model in this work
\begin{equation}
 \Gamma^{\rm QGP}_\gamma=\Gamma^{\rm AMY}_\gamma+\Delta\Gamma^{\rm pair}_\gamma,
 \label{eq:additive}
\end{equation}
where $\Gamma^{\rm AMY}_\gamma$ is the complete leading-order equilibrium rate of Arnold, Moore, and Yaffe, including the collinear bremsstrahlung and inelastic annihilation processes~\cite{Arnold:2001ms}. 
Both terms use the same flavor content ($u$, $d$, $s$), coupling prefactor $\kappa=16\pi^2\alpha_{\rm em}\alpha_s$, phase-space normalization, and units.

Our calculation has four steps.
First, we define the connected pair correlation by its vanishing one-particle marginals and build a library of its local photon-spectrum responses.
Second, we fold these responses over hydrodynamic profiles to obtain two quantities: $Y_0$, the azimuthally integrated photon yield, and $Y_2$, the second-harmonic moment, with $v_2 = Y_2 / Y_0$.
We add fixed prompt, pre-equilibrium, and thermal photon contributions and compare with the measurements in their published bins.
Third, we cast the reconstruction as a physics-constrained regression task. We select four response features from the candidate library and learn their coefficients separately for each collision class using the joint photon-yield and $v_2$ data.
Finally, we separate the correlation-induced changes in the yield and second harmonic and examine the response rank and local positivity.

Figure~\ref{fig:atlas} shows the response library.
Panel (a) shows the sign and shape of $79$ stored responses and of the separately stored factorized reference without correlation correction, labeled $F$, on a common grid $x \equiv k\cdot u/T \in [0.01, 15]$; each row is scaled by its own $\max_x|R_a|$.
Panel (b) shows the stored amplitudes, which span a factor of $4.8\times10^5$ between modes, and the number of sign changes, which ranges from $2$ to $8$ across the non-reference responses.
Every non-reference response changes sign.
Each mode therefore gives a signed correction to the emission rate.
Of the $79$ non-reference modes, $64$ have $\ell\ge1$ and nine have $\ell=0$ with $n,m\ge1$; all $73$ satisfy the zero-marginal conditions and form the candidate pool of the fit.
The remaining six $\ell=0$ modes do not satisfy these conditions and are excluded from the pair-correlation inversion.
Panel (c) shows one representative shape for each $\ell=0,\ldots,4$, selected among modes with zero marginals.
The stored amplitudes in panel (b) are calculated before fitting the mode coefficients and are not physical mode weights.
The figure illustrates the signs and shapes of the response modes.

\subsection{Hydrodynamic profiles and space-time convolution}
\label{sec:method_hydro}

The local rate is mapped to laboratory observables by a hydrodynamic profile. 
Each profile cell stores the proper time $\tau$, the transverse position $(x,y)$, the temperature $T$, and the transverse flow components $(v_x,v_y)$. Under longitudinal boost invariance the flow velocity is $u^\mu=\gamma_\perp(\cosh\eta_s,v_x,v_y,\sinh\eta_s)$ with $\gamma_\perp=(1-v_x^2-v_y^2)^{-1/2}$, and the four-volume in Milne coordinates is $\dd^4X=\tau\,\dd\tau\,\dd x\,\dd y\,\dd\eta_s$. A photon with transverse momentum $p_T$, rapidity $y$, and azimuth $\phi_p$ has the local energy
\begin{equation}
 E^*=k\!\cdot\!u=\gamma_\perp p_T\big[\cosh(y-\eta_s)-v_x\cos\phi_p-v_y\sin\phi_p\big].
 \label{eq:doppler}
\end{equation}
The QGP contribution at midrapidity follows from folding the local rate over all cells above the switching temperature,
\begin{equation}
 \frac{\dd N^{\rm QGP}_\gamma}{p_T\dd p_T\,\dd y\,\dd\phi_p}
 =\int_{T(X)>T_{\rm sw}}\dd^4X\;\Gamma^{\rm QGP}_\gamma\big(E^*=k\!\cdot\!u,T(X)\big),
 \label{eq:convolution}
\end{equation}
with $T_{\rm sw}=145~\mathrm{MeV}$. This switching temperature is the same as that in Ref.~\cite{Gale:2021emg}, which provides the baseline contributions from the other stages in our study. The flow term in Eq.~\eqref{eq:doppler} is the only source of azimuthal dependence: a locally isotropic emission becomes anisotropic in the laboratory because cells moving along the reaction plane emit harder photons in that direction.

We use the azimuthal moments
\begin{align}
 Y_n(p_T)=\int_0^{2\pi}\dd\phi_p\,\cos[n(\phi_p-\Psi_n)]\,
 \frac{\dd N_\gamma}{p_T\dd p_T\,\dd y\,\dd\phi_p},
 \label{eq:moments}
\end{align}
for $n=0,2$, averaged over each published $p_T$ bin $i$. Then $Y_{0,i}/(2\pi)$ is the invariant yield and $v_{2,i}=Y_{2,i}/Y_{0,i}$. Both $Y_0$ and $Y_2$ depend affinely on the coefficients of Eq.~\eqref{eq:local_increment}, whereas $v_2$ is a ratio. This is why the inverse problem below is formulated for the moments, not for $v_2$ directly.

The profiles used in this work are VISHydro~\cite{Shen:2014vra} profile with MCGlb initial condition for the four collision classes, with initial time $\tau_0=0.6~\mathrm{fm}/c$ and grid spacings $\Delta\tau=0.1~\mathrm{fm}/c$ and $\Delta x=\Delta y=0.5~\mathrm{fm}$. See Ref.~\cite{Shi:2018izg} for more details of the medium profile in use. 
The convolution keeps cells with $T>145~\mathrm{MeV}$, weights each cell by $\tau\Delta\tau \Delta x \Delta y$. 
These profiles are not identical to the IP-Glasma+K{\o}MP{\o}ST+\textsc{music} evolution of Ref.~\cite{Gale:2021emg}, but they are very similar.

\subsection{Other direct-photon sources}
\label{sec:method_sources}

Direct photons are photons not produced by hadron decays. We write their moments as
\begin{equation}
 Y^{\rm tot}_n=\underbrace{P_n+E_n+H_n}_{B_n}+\,C_n,\qquad
 C_n=\sum_\alpha\psi_\alpha\,Y_{n,\alpha}.
 \label{eq:sources}
\end{equation}
Here $P_n$, $E_n$, and $H_n$ are the prompt, pre-equilibrium thermal, and hydrodynamic thermal contributions, respectively, taken from the published source files of Ref.~\cite{Gale:2021emg}.
Photons produced during the prompt stage have no elliptic flow, so their $v_2$ vanishes, i.e., $P_2 = 0$.
We average these ``three'' contributions over the experimental bins in the same way as $C_n$ and use their sum $B_n$ as the internal fixed baseline.
The correlation increment $C_n$ is obtained by folding the local rate correction of Eq.~\eqref{eq:local_increment} over the VISHydro profiles.
Here $\alpha=(A,n,m,\ell)$ labels the fitted coefficients, and $Y_{n,\alpha}$ is the resulting moment at unit amplitude.

Three features of this construction limit its interpretation. 
First, $B_n$ and $C_n$ come from similar but not identical hydrodynamic evolutions.
Second, the AMY rate of Eq.~\eqref{eq:additive} is not added to the observables. The published thermal term $H_n$, which carries a slight viscous correction, plays its role, while $\Gamma^{\rm AMY}_\gamma$ enters only through the positivity condition on the local rate (Sec.~\ref{sec:results}). We use $H_n$ rather than the AMY rate in order to better assess the enhancement relative to the fit results of existing models.
Third, the chemical-equilibration suppression studied in Ref.~\cite{Gale:2021emg} is not applied to $C_n$. 

\subsection{Experimental data products}
\label{sec:method_data}

\begin{table*}
 \caption{Experimental inputs and their use. The centrality columns give the number of fitted bins or points out of all published ones. $^{a}$No baseline $v_2$ is available in these bins. $^{b}$Compared with the model at the published $p_T$ points; not fitted (Sec.~\ref{sec:method_data}). $^{c}$Two $95\%$ CL upper limits (UL) in the $20-40\%$ class, used as one-sided constraints.}
 \label{tab:data}
 \begin{ruledtabular}
 \begin{tabular}{lllccl}
 System & Observable & Measurement & $0-20\%$ & $20-40\%$ & Not fitted \\
 \hline
 Au+Au, $200~\mathrm{GeV}$ & $v_2$ & PHENIX, conversion~\cite{PHENIX:2015igl} & $10/11$ & $10/11$ & $3.5--5.0~\mathrm{GeV}${}$^{a}$ \\
 & yield & PHENIX, excess${}+N_{\rm coll}\,pp$~\cite{PHENIX:2014nkk} & $11/11$ & $11/11$ & --- \\
 & yield & STAR, virtual photons~\cite{STAR:2016use} & $0/9$ & $0/9$ & all points$^{b}$ \\
 Pb+Pb, $2.76~\mathrm{TeV}$ & $v_2$ & ALICE~\cite{ALICE:2018dti} & $13/16$ & $13/16$ & $4.1--6.2~\mathrm{GeV}${}$^{a}$ \\
 & yield & ALICE~\cite{ALICE:2015xmh} & $20/20$ & $18/20$+$2$ UL$^{c}$ & --- \\
 \end{tabular}
 \end{ruledtabular}
\end{table*}

Table~\ref{tab:data} lists the measurements and their use. 
In Au+Au collisions we use the PHENIX conversion-method $v_2$~\cite{PHENIX:2015igl} and the PHENIX direct-photon yield, formed as the published excess spectrum plus the $N_{\rm coll}$-scaled $p+p$ reference~\cite{PHENIX:2014nkk}.
STAR reports its direct virtual-photon yield at discrete $p_T$ values but does not provide bin boundaries~\cite{STAR:2016use}.
We therefore exclude the STAR data from the fit and compare the model with them at the published $p_T$ values.
In Pb+Pb collisions we use the ALICE direct-photon $v_2$~\cite{ALICE:2018dti} and yield~\cite{ALICE:2015xmh}.

Every other comparison uses the published $p_T$ bins. The highest PHENIX $v_2$ bin and the three ALICE $v_2$ bins above $4.1~\mathrm{GeV}$ are not fitted, because the published source files used for $B_n$ give no $v_2$ there. The uncertainty of each point is the quadrature sum of the statistical and systematic uncertainties, or the published total when only a total is given, and it is symmetrized for the fit. The two lowest ALICE $20\%$--$40\%$ yield points are $95\%$ confidence-level upper limits; they enter as one-sided constraints; see Appendix~\ref{app:inverse} for the exact treatment.

\section{Results}
\label{sec:results}

\begin{figure*}[!t]
 \centering
 \includegraphics[width=5.9in]{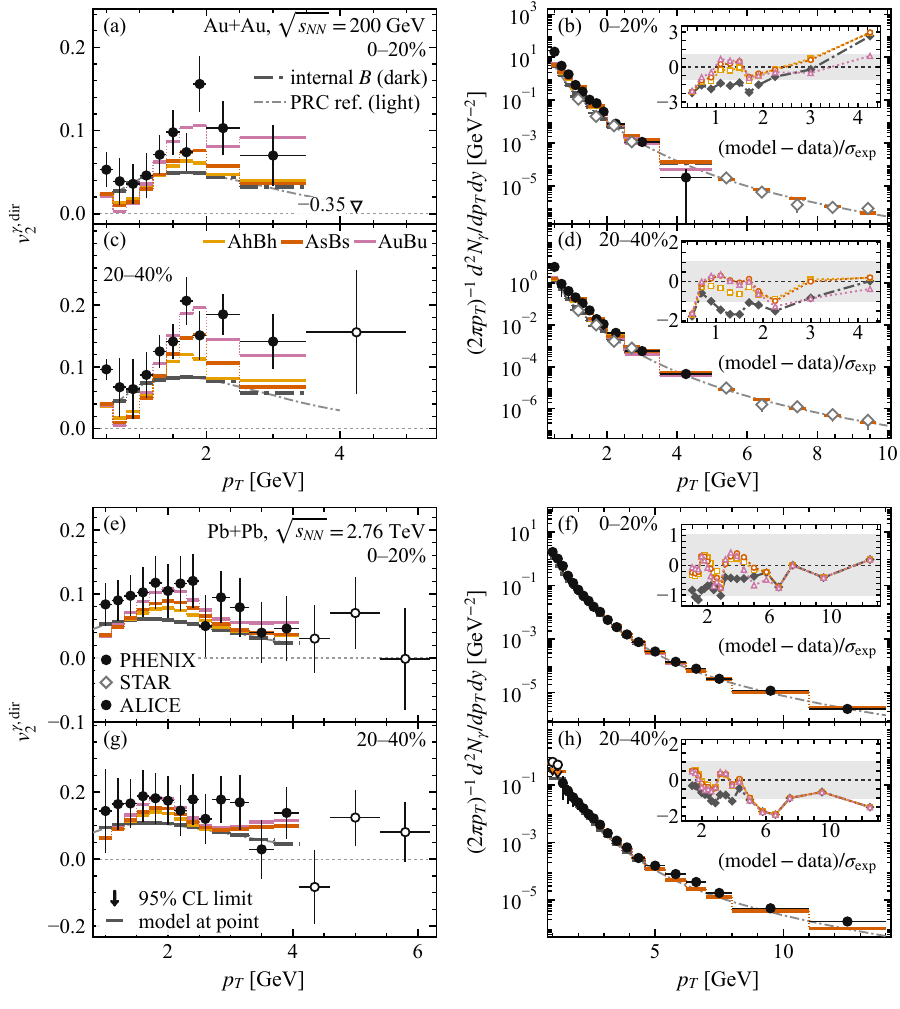}
 \caption{
 Direct-photon $v_2$ (left) and invariant yield (right) in Au+Au collisions at $\sNN=200~\mathrm{GeV}$, $0-20\%$ (a,b) and $20-40\%$ (c,d), and in Pb+Pb collisions at $\sNN=2.76~\mathrm{TeV}$, $0-20\%$ (e,f) and $20-40\%$ (g,h); the data are listed in Table~\ref{tab:data}.
 Vertical bars show the symmetrized total uncertainty used in the fit, horizontal bars the published bins.
 Black circles show PHENIX and ALICE measurements (filled when fitted, open otherwise), and gray open diamonds show STAR measurements at their published points; arrows are $95\%$ CL upper limits.
 Models are drawn bin by bin, with thin connectors between adjacent bins within the same binning scheme. 
 Short tick marks show the model values at the STAR points and at the ALICE limits. The thin light-gray dash-dotted curve is the public results of $B=P+E+H$ [Eq.~\eqref{eq:sources}] baseline given in Ref.~\cite{Gale:2021emg}, and the thick dark-gray dash-dotted curve is our treatment that recasts the continuous results therein into the same $p_T$ bins of the experiemental data.
 The three colored curves show the three $B+C$ reconstructions: gold denotes hard contraint with no violation, orange-red denotes soft contraint with $\tau_A=\tau_B=0.1$, and pink denotes unconstrained with $\tau_A=\tau_B=\infty$.
 The insets in panels (b,d,f,h) show $(Y_i^{\rm model}-Y_i^{\rm exp})/\sigma_i$ for the dark-gray baseline and three colored reconstructions in the fitted PHENIX/ALICE yield bins, with $\sigma_i$ the symmetrized experimental total uncertainty. 
 The light-gray band indicating $-1\le (Y_i^{\rm model}-Y_i^{\rm exp})/\sigma_i\le 1$.
 The open triangle in (a) marks a PHENIX data point outside the plotted range, $v_2=-0.35\pm0.48$, which was not fitted because no corresponding baseline data exist.
}
 \label{fig:joint}
\end{figure*}

To test whether the pair-correlation mechanism can improve both observables while satisfying the conditions for a physical local distribution and emission rate, Fig.~\ref{fig:joint} compares the measured direct-photon yields and $v_2$ with three reconstructions in each of the four collision classes.
The two gray styles show Ref.~\cite{Gale:2021emg} in different forms: the thin light-gray dash-dotted curve reproduces its published total calculation, while the thick dark-gray bin segments show the corresponding values in the experimental bins.
The binned values define the internal fixed baseline $B=P+E+H$ of Eq.~\eqref{eq:sources}.
We use this baseline to see how much the correlation term improves agreement with the experimental data.

Each colored curve is an in-sample fit using the same four response columns, with coefficients fitted separately for each class.
For each class we fit the yield and flow together using the following joint loss function
\begin{align}
\begin{split}
 \chi^2(\boldsymbol\psi)={}&\sum_{i\in Y}\Big(\frac{Y_i(\boldsymbol\psi)-Y^{\rm exp}_i}{\sigma_i}\Big)^2\\
 &+\sum_{i\in v_2}\Big(\frac{Y_{2,i}(\boldsymbol\psi)/Y_{0,i}(\boldsymbol\psi)-v^{\rm exp}_{2,i}}{\sigma_i}\Big)^2,
\end{split}
\label{eq:chi2_joint_v2}
\end{align}
so each fitted bin is weighted by its own uncertainty.
The two ALICE yield upper limits are included as one-sided constraints.
The resulting $\chi^2(\boldsymbol\psi)$ measures fit quality for the yield and flow data.

The three colored curves in Fig.~\ref{fig:joint} show different treatments of the following two physical constraints.
Constraint A requires a nonnegative total local photon rate, $r_A\equiv\Gamma^{\rm QGP}_\gamma/\Gamma^{\rm AMY}_\gamma\ge0$ for $0.31\le x\le15$ in cells with $T>T_{\rm sw}$.
Constraint B requires a nonnegative two-particle distribution, $r_B\equiv\mathcal F_{ab}/f_af_b\ge0$ for $\xi_a,\xi_b\le12$ and all $z_{ab}$ in those cells.
For a physical description, both constraints must be satisfied everywhere.
Given the limitations of our construction discussed in the preceding section, we also test a soft-constraint fit that permits negative values but penalizes their weighted mean-square violation.
The three fits are referred to as ``hard'', ``soft'', and ``unconstrained''.
The hard-constraint fit enforces both constraints on the test grids.
The soft-constraint fit, used as the main fit, adds $P=\tau^{-2}\langle\min(0,r)^2\rangle$ to Eq.~\eqref{eq:chi2_joint_v2} for each constraint, with the chosen penalty parameters $\tau_A=\tau_B=0.1$.
At $\tau=0.1$, a root-mean-square violation of $10\%$ costs one unit of $\chi^2$.
Both soft penalties give equal weight to hot cells.
The A penalty weights the tested $x$ points equally because it evaluates the local emission rate at each point, while the B penalty uses the thermal two-particle phase-space weight because it evaluates the two-particle distribution.
The unconstrained fit sets $\tau_A=\tau_B=\infty$ and leaves both constraints unenforced.
To keep $v_2=Y_2/Y_0$ numerically stable, we impose $Y_0^{\rm tot}\ge0.02B_0$ in the evaluated bins and points.
This numerical floor is inactive in all three reported fits.

Figure~\ref{fig:joint} shows how the joint yield and $v_2$ fits with a two-particle correlation term compare with the internal baseline and experimental data.
In the $v_2$ panels (a,c,e,g), the dark-gray baseline rises toward intermediate $p_T$ and then falls, but lies below nearly all measurements.
The colored reconstructions retain this broad momentum dependence, with a shallow low-$p_T$ dip in Au+Au before they rise toward a broad shoulder.
They tend to flatten toward the upper end of the modeled range and generally lie above the baseline at intermediate and higher $p_T$.
At the lowest momenta, however, their deviations from the baseline are small and can have either sign.

We next compare the three colored reconstructions.
The gold hard-constraint curve rises above the baseline around the intermediate-$p_T$ shoulder, whereas the orange-red soft-constraint curve develops a higher shoulder and the pink unconstrained curve gives the largest and broadest enhancement.
The three curves separate most clearly around the intermediate-$p_T$ shoulder.
In these fits, their separation shows how strongly the positivity constraints limit the flow added by the correlation term.

\begin{figure*}[!t]
 \centering
 \includegraphics[width=6.0in]{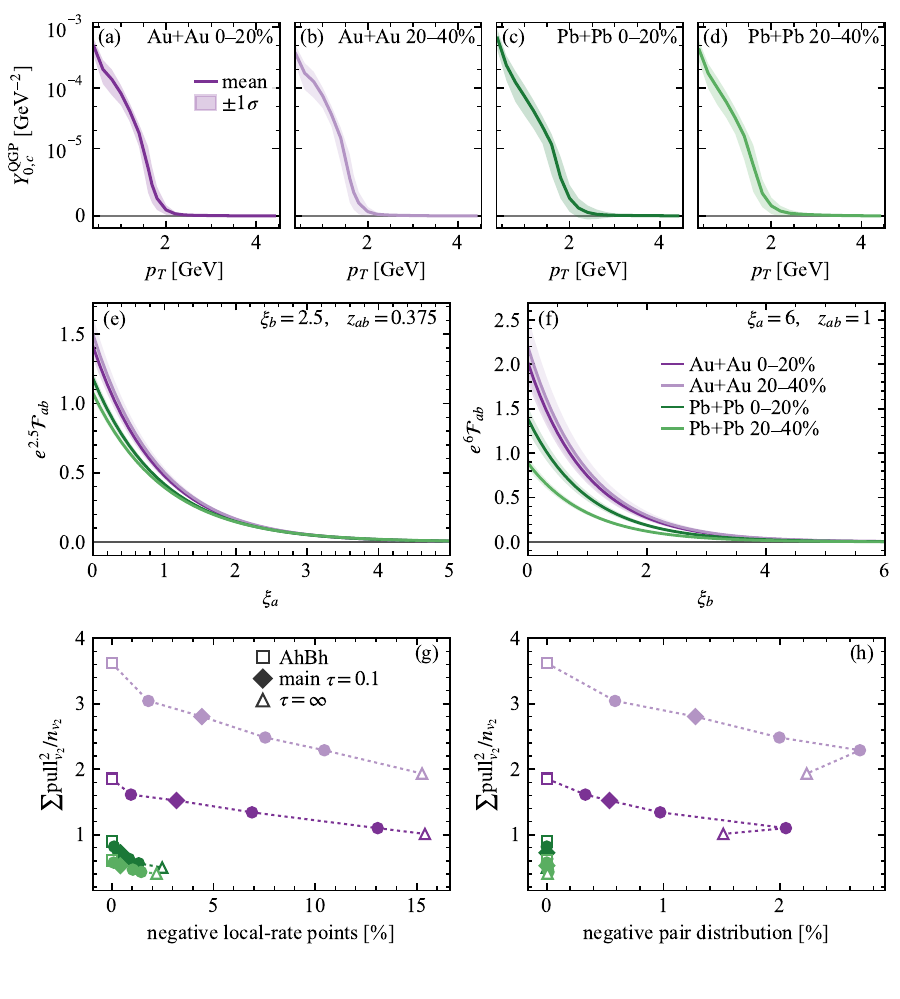}
 \caption{
 Diagnostics of the physical constraints.
 Panels (a)--(d) show the $p_T$ dependence of the thermal-photon yield from local fluid cells in Au+Au $0-20\%$, Au+Au $20-40\%$, Pb+Pb $0-20\%$, and Pb+Pb $20-40\%$, respectively.
 The solid curves are arithmetic means across all hot cells, and the shaded regions show one population standard deviation among cells.
 The vertical axes use a symmetric logarithmic scale that is linear for $|Y_{0,c}^{\rm QGP}|\le10^{-5}~\mathrm{GeV}^{-2}$ and displays negative values.
 Panels (e,f) show two cuts of the main-fit local pair distribution $\mathcal F_{ab}$.
 Panel (e) varies $\xi_a$ at fixed $\xi_b=2.5$ and $z_{ab}=0.375$, with $e^{2.5}\mathcal F_{ab}$ on the vertical axis.
 Panel (f) varies $\xi_b$ at fixed $\xi_a=6$ and $z_{ab}=1$, with $e^{6}\mathcal F_{ab}$ on the vertical axis.
 The solid curves are four-volume-weighted means across hot cells, and the shaded regions show one standard deviation.
 The colors in (e)--(h) follow the collision-class order of (a)--(d).
 Panels (g,h) show the summed squared $v_2$ pulls per fitted bin on the vertical axis.
 Their horizontal axes show the percentage of negative local-rate test points in (g) and the percentage of negative pair-distribution values in (h).
 For each class, a dotted line joins the hard-constraint fit and the fits with $\tau_A=\tau_B=\tau$ at $\tau=0.03$, $0.1$, $0.3$, $1$, and $\infty$, in that order.
 Open squares mark the hard-constraint fit (A and B hard), filled diamonds the main soft-constraint fit ($\tau=0.1$), open triangles the unconstrained fit ($\tau=\infty$), and filled circles the other soft-constraint fits.
 }
 \label{fig:cells_pair}
\end{figure*}

The comparison with the measurements shows how these curve differences affect the $v_2$ fit.
Most measured $v_2$ points remain above the baseline and the hard-constraint curve.
Around the broad measured enhancement, the soft-constraint fit moves closer to the data, and the unconstrained fit often comes closer still or reaches the measured central values.
The gap persists at low $p_T$, especially in Au+Au, and the unconstrained curve is not closest to every point at higher $p_T$.
Across these three reconstructions, the in-sample joint fit describes $v_2$ better as the constraints are relaxed.
The cost of this better fit is that negative local photon rates and negative pair distributions become more frequent, as shown later in Fig.~\ref{fig:cells_pair}(g,h).

The yield fits improve more clearly than the $v_2$ fits.
In the logarithmic main panels of Fig.~\ref{fig:joint}(b,d,f,h), the baseline appears close to the measurements, but the insets reveal a mainly negative offset, with several Au+Au points outside the $1\sigma$ band.
Adding the pair correlation raises the yield toward the fitted data over low and intermediate $p_T$.
This improvement is already clear for the hard-constraint fit, which enforces both positivity constraints.
Below about $6\,\mathrm{GeV}$, nearly all Pb+Pb and Au+Au $20-40\%$ points in the hard-constraint fit lie within the $1\sigma$ band, and Au+Au $0-20\%$ also improves, although some points there remain outside the band.
For all three Pb+Pb fits and for the unconstrained fit in Au+Au, the residuals have both signs, although the hard and soft-constraint fits in Au+Au remain mostly below PHENIX.
At high $p_T$ in Pb+Pb, the reconstructions return to the baseline because the present pair-response library sets the correction to zero beyond $x=15$.

The remaining low-$p_T$ differences in yield and $v_2$ motivate a closer look at the pair-correlation response.
Here this response includes only $2\to2$ Compton and annihilation processes in the classical-statistics, leading-logarithmic limit~\cite{Lu:2026nmp}.
At complete leading order, the equilibrium rate also includes near-collinear bremsstrahlung and inelastic pair annihilation, with Landau--Pomeranchuk--Migdal suppression~\cite{Arnold:2001ms}.
These channels enter the AMY rate used for the local positivity condition, but their response to the pair correlation is absent from the fitted correction.
The classical limit also drops the final-state occupation factors in Eq.~\eqref{eq:statistical_factor}.
A correlation-induced correction from these missing terms could change $Y_0$ and $Y_2$ differently after integration over the medium and thus affect $v_2=Y_2/Y_0$ and yield.
Extending the pair response to complete leading order, and then assessing next-to-leading-order corrections, would test whether these approximations contribute to the remaining low-$p_T$ differences.

\begin{figure*}
 \centering
 \includegraphics[width=6.0in]{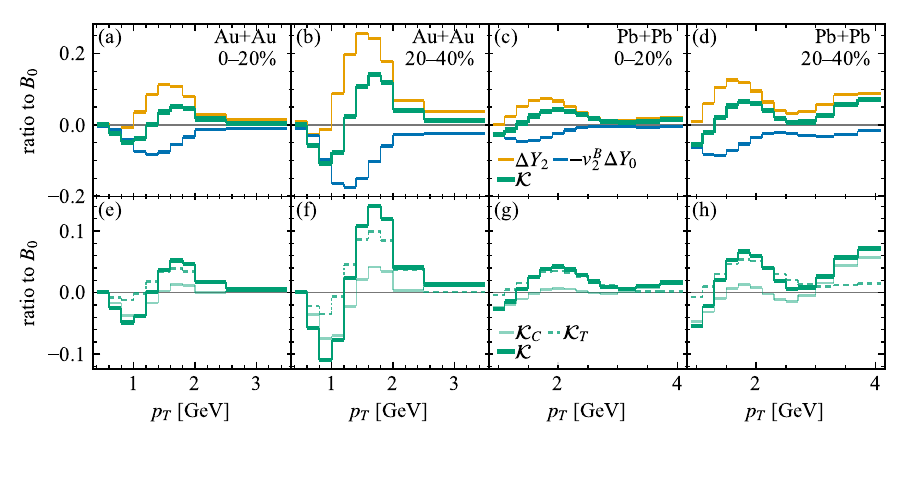}
 \caption{
 Decomposition for the main fit.
 From left to right, the columns correspond to Au+Au $0-20\%$, Au+Au $20-40\%$, Pb+Pb $0-20\%$, and Pb+Pb $20-40\%$, rspectively.
 Upper panels (a)--(d) show the full-fit $\Delta Y_2/B_0$ in orange-yellow, $-v_2^B\Delta Y_0/B_0$ in blue, and their sum $\mathcal K/B_0$ in green in the fitted $v_2$ bins.
 The green $\mathcal K/B_0$ line is thicker than the orange-yellow and blue lines.
 Lower panels (e)--(h) show $\mathcal K/B_0$ as an opaque dark green line, its constant-column part $\mathcal K_C/B_0$ as a semi-transparent solid line, and its temperature-column part $\mathcal K_T/B_0$ as a semi-transparent dashed line.
 }
 \label{fig:moments}
\end{figure*}

In this section, we examine the origin of the change in $v_2$.
Relative to the baseline $v_2^B\equiv B_2/B_0$, the change in $v_2$ is exactly
\begin{equation}
 v_2^{\rm tot}-v_2^B=\frac{\mathcal K}{B_0+\Delta Y_0},\qquad
 \mathcal K\equiv\Delta Y_2-v_2^B\Delta Y_0 .
 \label{eq:leverage_v2}
\end{equation}
For a positive total yield, $v_2$ rises if and only if $\mathcal K>0$.
The quantity $\mathcal K$ has two terms.
The first term is the added second harmonic.
The second term, $-v_2^B\Delta Y_0$, is the negative of the second harmonic expected if the added yield had the baseline anisotropy.
If the correlation added only a positive second harmonic without changing the yield, then $\Delta Y_0=0$ and $\mathcal K=\Delta Y_2>0$.
If the correlation added only a positive yield without changing the second harmonic, then $\Delta Y_2=0$ and $\mathcal K=-v_2^B\Delta Y_0<0$ for $v_2^B>0$.

The upper panels of Fig.~\ref{fig:moments} show $-v_2^B\Delta Y_0/B_0$ in blue and the second-harmonic increment $\Delta Y_2/B_0$ in orange-yellow for the main fit.
The green line shows the full $v_2$ change $\mathcal K/B_0$, and the green line is the sum of the orange-yellow and blue lines.
Around the intermediate-$p_T$ enhancement, $v_2^B\Delta Y_0$ and $\Delta Y_2$ are often of the same order, and the absolute value of the yield term is smaller where $\mathcal K>0$.
Near the shoulder, the pair correlation adds a second harmonic, but the accompanying increase in yield offsets part of its effect on $v_2$.
The resulting rise in $v_2$ is smaller than one would infer from $\Delta Y_2$ alone.
In some low-$p_T$ bins, $\mathcal K$ is negative, so the correlation does not raise $v_2$ in every bin.

We next isolate the contribution of the two temperature-dependent columns while keeping all main-fit coefficients fixed.
Let $\Delta Y_n^{(C)}$ denote the contributions from the two constant-feature columns, with $\mathcal K_C\equiv\Delta Y_2^{(C)}-v_2^B\Delta Y_0^{(C)}$.
The temperature-column contribution is then
\begin{align}
 \Delta Y_n^{(T)}&\equiv\Delta Y_n-\Delta Y_n^{(C)},\\
 \mathcal K_T&\equiv\mathcal K-\mathcal K_C
 =\Delta Y_2^{(T)}-v_2^B\Delta Y_0^{(T)}.
\end{align}
In the lower panels of Fig.~\ref{fig:moments}, the  semi-transparent solid curves show $\mathcal K_C/B_0$, and the semi-transparent dashed curves show $\mathcal K_T/B_0$ obtained by subtracting them from the opaque solid curves.
This is a decomposition of the main fit rather than a separate fit with only the temperature columns.
At the intermediate-$p_T$ shoulder in these panels, $\mathcal K_T$ supplies most of the positive remainder in all four collision classes, while the constant columns contribute more in the highest fitted bins.

This comparison suggests that the temperature-dependent correlation helps produce a positive $\mathcal K$ by changing the relative contributions of hotter and cooler regions.
Without the pair correlation, the temperature weighting of each fluid cell's thermal emission is constrained by the local one-particle distributions and the equation of state.
The connected pair correlation has zero one-particle marginals [Eq.~\eqref{eq:zero_marginals}], so it leaves the one-particle distributions unchanged.
Its amplitude can depend on $q_T=\ln[T/(0.2~\mathrm{GeV})]$, giving the correlation correction an additional temperature-dependent weight across fluid cells.

The pattern is less uniform at low $p_T$, which may partly reflect the leading-logarithmic approximation used for the pair-correlation response, as discussed above.

\begin{figure*}[!t]
 \centering
 \includegraphics[width=6.0in]{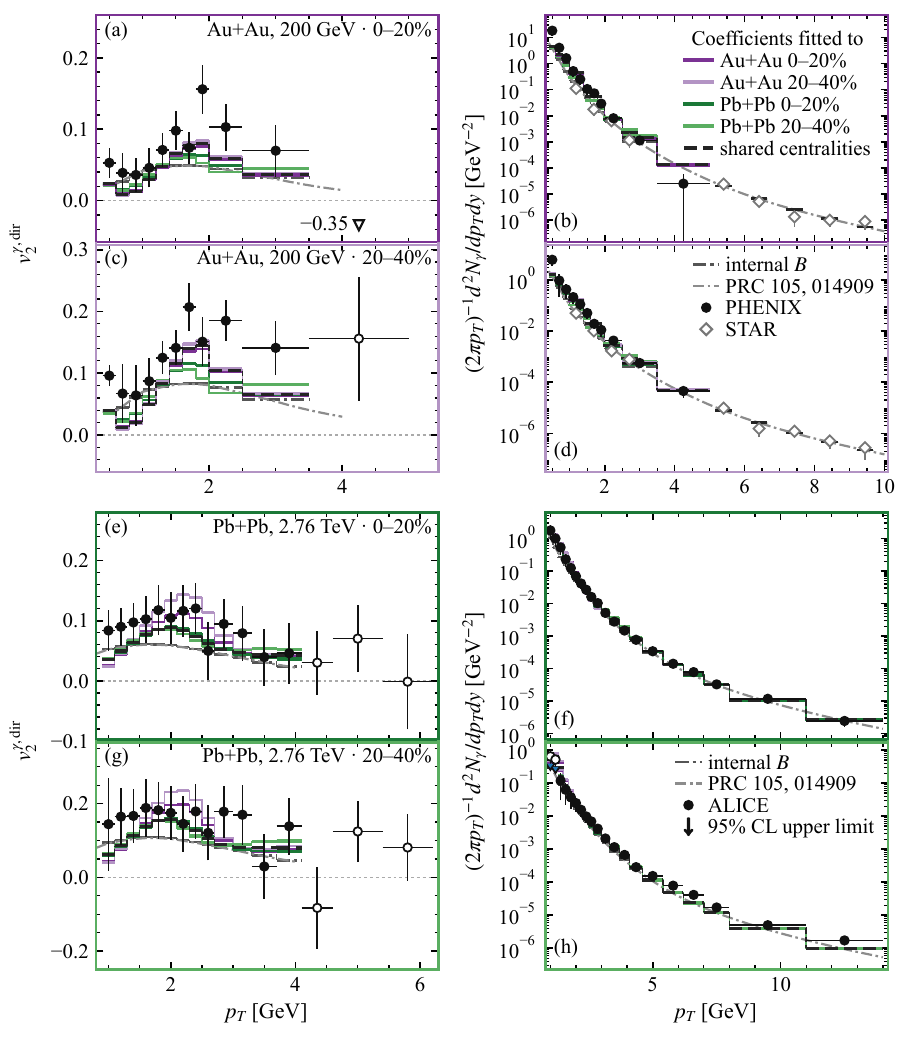}
 \caption{
 Transfer of the four class-specific main-fit coefficient vectors and fits shared across the two centralities of each collision system.
 Panels (a,b), (c,d), (e,f), and (g,h) target Au+Au $0-20\%$, Au+Au $20-40\%$, Pb+Pb $0-20\%$, and Pb+Pb $20-40\%$, respectively. The left panel of each pair shows direct-photon $v_2$ and the right panel the invariant yield.
 The colored border of each panel follows the same color in Fig.~\ref{fig:cells_pair}(a)--(d) and identifies its target collision class; curve color identifies the source class of the transferred coefficients.
 In each target pair, the heavier curve is the target's own fit.
 The dark dashed curve in each panel uses the coefficient vector fitted jointly to the two centralities of that collision system.
 The two gray dash-dotted curves show the same calculation from Ref.~\cite{Gale:2021emg}: the thick dark-gray curve shows its binned form, used as the internal baseline $B$, while the thin light-gray curve shows its published continuous form.
 Filled circles show the fitted PHENIX or ALICE measurements with uncertainties, open $v_2$ circles denote published bins excluded from the fit, and open diamonds show STAR yields at their published $p_T$ values.
 Colored and dark short ticks at the STAR points and at the two ALICE $20-40\%$ upper-limit bins mark the class-specific and shared-fit model predictions at those points.
 Downward arrows indicate the experimental $95\%$ confidence-level upper limits.
 }
 \label{fig:shared}
\end{figure*}

\begin{figure*}
 \centering
 \includegraphics[width=6.0in]{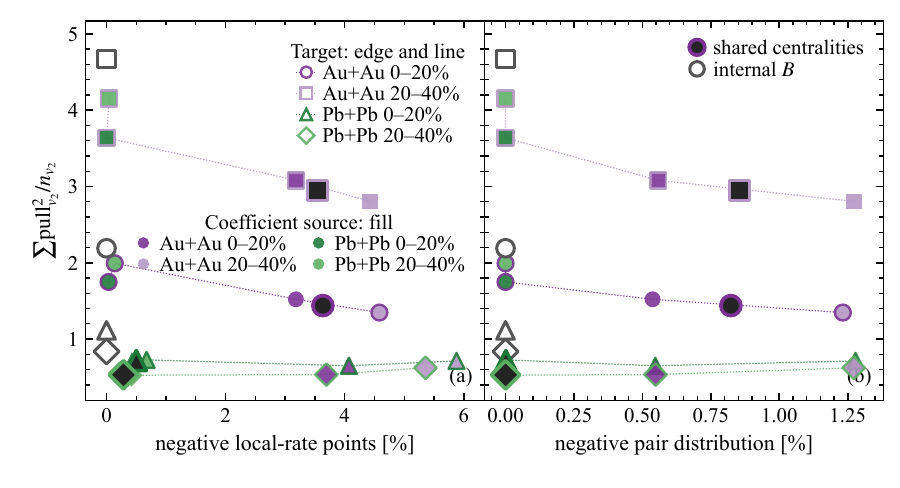}
 \caption{
 Fit quality and positivity diagnostics for transferred, shared, and baseline coefficients.
 Each of the four main-fit coefficient vectors is applied to every target class without refitting.
 Both panels show the summed squared $v_2$ pulls per fitted bin for the target class.
 Panel (a) gives the percentage of negative local-rate test points, while panel (b) gives the percentage of negative pair-distribution values.
 Marker shape identifies the target class: circles and squares represent Au+Au $0-20\%$ and $20-40\%$, and triangles and diamonds represent Pb+Pb $0-20\%$ and $20-40\%$, respectively.
 For the transferred coefficients, outline color identifies the target class and fill color identifies the source class.
 Matching fill and outline colors identify a target class's own fit.
 The larger dark-filled markers with target-colored outlines show the coefficients fitted jointly to the two centralities of the corresponding collision system.
 Gray-outlined open markers show the internal baseline $B$, for which both negative fractions are zero.
 Dotted lines connect the four source vectors for each target as a visual guide and do not represent a parameter scan.
 }
 \label{fig:transfer_tradeoff}
\end{figure*}

\begin{figure}
 \centering
 \includegraphics[width=3.0in]{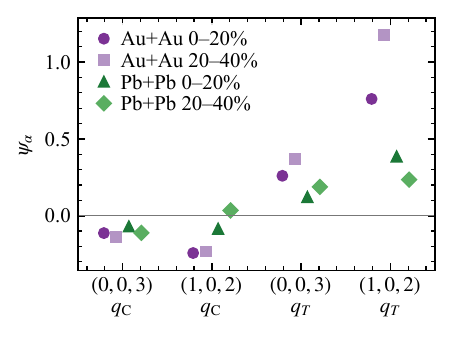}
 \caption{Coefficients of the four dominating modes $\psi_\alpha \equiv \psi_{A,nlm}$ with $A = C$ or $T$ and $(n,l,m) = (0,0,3)$ or $(1,0,2)$. Different symbols represent different collision systems, and they have been shifted hotizontally for better visibility.}
 \label{fig:main_fit_coefficients}
\end{figure}

We now examine whether the reconstructed correlation satisfies the two physical positivity conditions introduced above, the result in shown in Figure~\ref{fig:cells_pair}.

Panels (a)--(d) show the four-volume-weighted laboratory-frame QGP yield contribution from individual hot cells as a function of $p_T$.
The solid curves are the means across cells, and the shaded regions show one standard deviation of the cell contributions.
The means remain positive and fall toward zero with increasing $p_T$, although a small fraction of cells contributes negatively at the shown momenta.

Panels (e,f) show two cuts of the local pair distribution~\eqref{eq:pair_decomposition}, $\mathcal F_{ab}$, from the main fit.
Panel (e) varies temperature-scaled-local-energy of particle $a$, $\xi_a$, at fixed $\xi_b=2.5$ and relative overlap $z_{ab}=0.375$, and displays $e^{\xi_b} \mathcal F_{ab}$.
Panel (f) varies $\xi_b$ at fixed $\xi_a=6$ and $z_{ab}=1$, and displays $e^{\xi_a} \mathcal F_{ab}$.
The solid curves are four-volume-weighted means across hot cells, and the shaded regions show one standard deviation.
The means and shaded bands remain positive over both displayed ranges.

Panels (a)--(d) and (e,f) show averages or selected cuts, which do not establish positivity over the full local domain.
The positivity conditions require a nonnegative local rate over the full range of $x$ and a nonnegative pair distribution over the full ranges of $\xi_a$, $\xi_b$, and $z_{ab}$.
These result are shown in panels (g,h) of Fig.~\ref{fig:cells_pair}, which quantify the relation between the $v_2$ fit quality and violations of local positivity.
Both panels plot $\sum_{i\in v_2}\mathrm{pull}_{v_2,i}^2/n_{v_2}$, the mean squared pull over the fitted $v_2$ bins, on the vertical axis.
Here $\mathrm{pull}_{v_2,i}=(v_{2,i}^{\rm model}-v_{2,i}^{\rm exp})/\sigma_i$, $\sigma_i$ is the symmetrized total experimental uncertainty, and $n_{v_2}$ is the number of fitted $v_2$ bins.
The horizontal axes give the percentage of negative local-rate points in (g) and the thermal-pair-weighted percentage of negative pair-distribution values in (h).
Each color identifies a collision class, and each dotted line connects its fits as the constraints are relaxed.
Open squares mark the hard-constraint fit, filled diamonds the main soft-constraint fit with $\tau_A=\tau_B=0.1$, and open triangles the unconstrained fit.
Filled circles mark the other soft-constraint fits, with $\tau_A=\tau_B$ set to $0.03$, $0.3$, or $1$.
From the hard-constraint fit through the main soft-constraint fit to the unconstrained fit, the mean squared $v_2$ pull decreases in all four collision classes, while both negative fractions increase.

This trade-off motivates a future calculation of the baseline and pair-correlation response with the same hydrodynamic profile and extending the latter to complete leading order.
These calculations would test whether the $v_2$ fit can improve while both positivity conditions are satisfied.

We next test whether or not our fitted coefficients are generalizable, i.e., whether those fitted for one collision class can be tranferred and describe another.
Figure~\ref{fig:shared} compares the transferred yield and $v_2$ curves with the target-class data.
Purple and light purple curves use coefficients fitted to Au+Au $0-20\%$ and $20-40\%$, while dark and light green curves use coefficients fitted to the corresponding Pb+Pb classes.
The border color identifies the target class, so the matching, heavier curve shows its own fit and the other colored curves show transfers.
Figure~\ref{fig:transfer_tradeoff} shows the $v_2$ fit quality and positivity diagnostics for these transfers.
The dark dashed curves in Fig.~\ref{fig:shared} and the dark-filled markers in Fig.~\ref{fig:transfer_tradeoff} show two additional fits, one for each collision system, in which both centralities share one coefficient vector.
The thick dark-gray dash-dotted curves in Fig.~\ref{fig:shared} and the open gray-outlined markers at zero negative fraction in Fig.~\ref{fig:transfer_tradeoff} show the internal baseline $B$.

The yield curves change less with the coefficient source than the $v_2$ curves.
Within each collision energy, coefficients from one centrality retain most of the joint yield and $v_2$ improvement over the baseline in the other centrality.
The transferred $v_2$ pulls per fitted bin also remain close to those of the target class's own fit.
Summed over the fitted yield and $v_2$ bins of both centralities, the squared pulls rise from $64.7$ for the two separate Au+Au fits to $66.0$ for their shared fit, and from $31.9$ to $33.4$ for Pb+Pb.

Across collision energies, the transfer is asymmetric.
Compared with each Pb+Pb target's own fit, Au+Au coefficients can give similar or smaller $v_2$ pulls, but larger negative fractions for both the local rate and the pair distribution.
Conversely, compared with each Au+Au target's own fit, Pb+Pb coefficients reduce both negative fractions, but give larger $v_2$ pulls and retain less of the improvement in the joint fit relative to the baseline.
Figure~\ref{fig:main_fit_coefficients} shows the fitted coefficients $\psi_{{\rm C},nm\ell}$ and $\psi_{T,nm\ell}$ for the two selected momentum modes, $(n,m,\ell)=(0,0,3)$ and $(1,0,2)$.
For both modes, the Au+Au fits give larger values of $\psi_{T,nm\ell}$ than the Pb+Pb fits, so the corresponding mode amplitudes increase more rapidly with $\ln T$.
Applying the Au+Au coefficients to the hotter Pb+Pb profiles thus extends this temperature dependence beyond the range in the Au+Au profiles, which may contribute to the increased positivity violations.
Applying the Pb+Pb coefficients to the Au+Au profiles weakens the temperature-dependent reweighting between hotter and cooler fluid cells, which may worsen the $v_2$ fit.

Within each collision energy, the centrality-transfer and shared-fit tests support using one coefficient vector for the two centralities studied.
Across energies, however, the transfer involves a trade-off between $v_2$ fit quality and local positivity.
Within the selected basis, the present results do not establish a common coefficient vector that describes both collision energies while satisfying both local positivity conditions.

\section{Conclusions}
\label{sec:conclusion}

With a forward model in which a connected two-particle correlation of the emitting partons changes the local $2\to2$ photon rate developed by the Authors in Ref.~\cite{Lu:2026nmp}, we ask whether there exists such a correlation can describe direct-photon yields and elliptic flow together with physically meaningful parameters. 
Connected modes are defined by zero marginals which leave the one-particle distributions unchanged, allowing us to isolate the pair-correlation response.
We folded this response over hydrodynamic profiles, added it to a fixed direct-photon baseline, and fitted the measured yields and $v_2$ while varying the enforcement of local nonnegativity.

The fitted correlation response changes both the yield and $v_2$ appreciably relative to the baseline.
The correlation response improves the fitted PHENIX and ALICE yields relative to the baseline even with both local-positivity conditions enforced, although it also increases the tension with the unfitted STAR yields in Au+Au.
Under these hard conditions, the $v_2$ fit also improves, though less than when the conditions are relaxed.
Relaxing the conditions improves the in-sample $v_2$ fit more strongly, but in the main fit $0.4\%--4.4\%$ of tested local-rate points and at most $1.3\%$ of the weighted pair domain are negative.
In the unconstrained fits, further improvement accompanies negative local rates at $2\%--15\%$ of tested points.

Within the selected basis, the two centralities at each collision energy are consistent with a shared coefficient vector, whereas transfer between energies is asymmetric.

The improved agreement may reflect a physical pair-correlation mechanism, although the use of different hydrodynamic evolutions for the baseline and response and the leading-logarithmic approximation for the response limit the conclusions that can be drawn.
To address these limitations, we will compute the baseline and pair-correlation response on the same hydrodynamic evolution with a consistent coupling.
We will also extend the pair-correlation response to complete leading order and then assess next-to-leading-order corrections.
These calculations will test whether a fit to both observables can satisfy both local positivity conditions, and they pave the way for a future study that incorporates connected correlations obtained by solving the transport equation under conditions relevant to heavy-ion collisions.

\section*{Acknowledgments} 
This work is supported by Yantai university under the grant 2226001.
This work is also supported by NSFC by grant No. 12575143 and by Tsinghua University under grant Nos. 04200500123, 531205006, 533305009. We also acknowledge the support by center of high performance computing, Tsinghua University.

\bibliography{biblio}

@article{Feinberg:1976ua,
    author = "Feinberg, E. L.",
    title = "{Direct Production of Photons and Dileptons in Thermodynamical Models of Multiple Hadron Production}",
    reportNumber = "CERN-TH-2156",
    journal = "Nuovo Cim. A",
    volume = "34",
    pages = "391",
    year = "1976"
}

@article{Shuryak:1978ij,
    author = "Shuryak, Edward V.",
    title = "{Quark-Gluon Plasma and Hadronic Production of Leptons, Photons and Psions}",
    reportNumber = "IYF-78-24",
    doi = "10.1016/0370-2693(78)90370-2",
    journal = "Phys. Lett. B",
    volume = "78",
    pages = "150",
    year = "1978"
}

@article{Kajantie:1981wg,
    author = "Kajantie, K. and Miettinen, H. I.",
    title = "{Temperature Measurement of Quark-Gluon Plasma Formed in High-Energy Nucleus-Nucleus Collisions}",
    reportNumber = "HU-TFT-81-7",
    doi = "10.1007/BF01548770",
    journal = "Z. Phys. C",
    volume = "9",
    pages = "341",
    year = "1981"
}

@article{McLerran:1984ay,
    author = "McLerran, Larry D. and Toimela, T.",
    title = "{Photon and Dilepton Emission from the Quark - Gluon Plasma: Some General Considerations}",
    reportNumber = "FERMILAB-PUB-84-084-T",
    doi = "10.1103/PhysRevD.31.545",
    journal = "Phys. Rev. D",
    volume = "31",
    pages = "545",
    year = "1985"
}

@article{Kajantie:1986dh,
    author = "Kajantie, K. and Kapusta, Joseph I. and McLerran, Larry D. and Mekjian, A.",
    title = "{Dilepton Emission and the QCD Phase Transition in Ultrarelativistic Nuclear Collisions}",
    reportNumber = "Print-86-0414 (MINNESOTA)",
    doi = "10.1103/PhysRevD.34.2746",
    journal = "Phys. Rev. D",
    volume = "34",
    pages = "2746",
    year = "1986"
}

@article{Tserruya:2009zt,
    author = "Tserruya, Itzhak",
    editor = "Stock, R.",
    title = "{Electromagnetic Probes}",
    eprint = "0903.0415",
    archivePrefix = "arXiv",
    primaryClass = "nucl-ex",
    reportNumber = "WIS-IT-2009-1",
    doi = "10.1007/978-3-642-01539-7_7",
    journal = "Landolt-Bornstein",
    volume = "23",
    pages = "176",
    year = "2010"
}

@inproceedings{Gale:2025ome,
    author = "Gale, Charles",
    title = "{Electromagnetic Radiation from High-Energy Nuclear Collisions}",
    eprint = "2502.13938",
    archivePrefix = "arXiv",
    primaryClass = "nucl-th",
    month = "2",
    year = "2025"
}

@article{Jackson:2025mtk,
    author = "Jackson, Greg",
    title = "{Theory overview on electroweak emission from heavy-ion collisions}",
    eprint = "2509.00262",
    archivePrefix = "arXiv",
    primaryClass = "hep-ph",
    doi = "10.1051/epjconf/202636401023",
    journal = "EPJ Web Conf.",
    volume = "364",
    pages = "01023",
    year = "2026"
}

@article{Du:2025ith,
    author = "Du, Lipei",
    title = "{Advancing multimessenger approaches in heavy-ion collisions: Insights from electromagnetic probes}",
    eprint = "2503.09045",
    archivePrefix = "arXiv",
    primaryClass = "nucl-th",
    doi = "10.1051/epjconf/202533901016",
    journal = "EPJ Web Conf.",
    volume = "339",
    pages = "01016",
    year = "2025"
}

@article{PHENIX:2008uif,
    author = "Adare, A. and others",
    collaboration = "PHENIX",
    title = "{Enhanced production of direct photons in Au+Au collisions at $\sqrt{s_{NN}}=200$ GeV and implications for the initial temperature}",
    eprint = "0804.4168",
    archivePrefix = "arXiv",
    primaryClass = "nucl-ex",
    doi = "10.1103/PhysRevLett.104.132301",
    journal = "Phys. Rev. Lett.",
    volume = "104",
    pages = "132301",
    year = "2010"
}

@article{PHENIX:2011oxq,
    author = "Adare, A. and others",
    collaboration = "PHENIX",
    title = "{Observation of direct-photon collective flow in $\sqrt{s_{NN}}=200$ GeV Au+Au collisions}",
    eprint = "1105.4126",
    archivePrefix = "arXiv",
    primaryClass = "nucl-ex",
    doi = "10.1103/PhysRevLett.109.122302",
    journal = "Phys. Rev. Lett.",
    volume = "109",
    pages = "122302",
    year = "2012"
}

@article{Liu:2012ax,
    author = "Liu, Fu-Ming and Liu, Sheng-Xu",
    title = "{Quark-gluon plasma formation time and direct photons from heavy ion collisions}",
    eprint = "1212.6587",
    archivePrefix = "arXiv",
    primaryClass = "nucl-th",
    doi = "10.1103/PhysRevC.89.034906",
    journal = "Phys. Rev. C",
    volume = "89",
    number = "3",
    pages = "034906",
    year = "2014"
}

@article{Rapp:2014hha,
    author = "Rapp, Ralf and van Hees, Hendrik",
    title = "{Thermal Dileptons as Fireball Thermometer and Chronometer}",
    eprint = "1411.4612",
    archivePrefix = "arXiv",
    primaryClass = "hep-ph",
    doi = "10.1016/j.physletb.2015.12.065",
    journal = "Phys. Lett. B",
    volume = "753",
    pages = "586--590",
    year = "2016"
}

@article{ALICE:2018dti,
    author = "Acharya, Shreyasi and others",
    collaboration = "ALICE",
    title = "{Direct photon elliptic flow in Pb-Pb collisions at $\sqrt{s_{\rm NN}}$ = 2.76 TeV}",
    eprint = "1805.04403",
    archivePrefix = "arXiv",
    primaryClass = "nucl-ex",
    reportNumber = "CERN-EP-2018-117",
    doi = "10.1016/j.physletb.2018.11.039",
    journal = "Phys. Lett. B",
    volume = "789",
    pages = "308--322",
    year = "2019"
}

@article{Dasgupta:2020orj,
    author = "Dasgupta, Pingal and Ma, Guo-Liang and Chatterjee, Rupa and Yan, Li and Zhang, Song and Ma, Yu-Gang",
    title = "{Thermal photons as a sensitive probe of $\alpha $-cluster in C + Au collisions at the BNL Relativistic Heavy Ion Collider}",
    eprint = "2007.09543",
    archivePrefix = "arXiv",
    primaryClass = "nucl-th",
    doi = "10.1140/epja/s10050-021-00441-8",
    journal = "Eur. Phys. J. A",
    volume = "57",
    number = "4",
    pages = "134",
    year = "2021"
}

@article{Sun:2023rhh,
    author = "Sun, Jing-An and Yan, Li",
    title = "{Estimating the magnetic field strength in heavy-ion collisions via direct photon elliptic flow}",
    eprint = "2311.03929",
    archivePrefix = "arXiv",
    primaryClass = "nucl-th",
    doi = "10.1103/PhysRevC.109.034917",
    journal = "Phys. Rev. C",
    volume = "109",
    number = "3",
    pages = "034917",
    year = "2024"
}

@article{Stankus:2005eq,
    author = "Stankus, P.",
    title = "{Direct photon production in relativistic heavy-ion collisions}",
    doi = "10.1146/annurev.nucl.53.041002.110533",
    journal = "Ann. Rev. Nucl. Part. Sci.",
    volume = "55",
    pages = "517--554",
    year = "2005"
}

@article{Liu:2007tw,
    author = "Liu, Fu-Ming and Werner, Klaus",
    title = "{A Systematic study on direct photon production from central heavy ion collisions}",
    eprint = "0712.3619",
    archivePrefix = "arXiv",
    primaryClass = "hep-ph",
    doi = "10.1088/0954-3899/36/3/035101",
    journal = "J. Phys. G",
    volume = "36",
    pages = "035101",
    year = "2009"
}

@article{Vitev:2008vk,
    author = "Vitev, Ivan and Zhang, Ben-Wei",
    title = "{A Systematic study of direct photon production in heavy ion collisions}",
    eprint = "0804.3805",
    archivePrefix = "arXiv",
    primaryClass = "hep-ph",
    doi = "10.1016/j.physletb.2008.10.019",
    journal = "Phys. Lett. B",
    volume = "669",
    pages = "337--344",
    year = "2008"
}

@article{David:2019wpt,
    author = "David, Gabor",
    title = "{Direct real photons in relativistic heavy ion collisions}",
    eprint = "1907.08893",
    archivePrefix = "arXiv",
    primaryClass = "nucl-ex",
    doi = "10.1088/1361-6633/ab6f57",
    journal = "Rept. Prog. Phys.",
    volume = "83",
    number = "4",
    pages = "046301",
    year = "2020"
}

@article{Owens:1986mp,
    author = "Owens, J. F.",
    title = "{Large Momentum Transfer Production of Direct Photons, Jets, and Particles}",
    reportNumber = "FSU-HEP-861021",
    doi = "10.1103/RevModPhys.59.465",
    journal = "Rev. Mod. Phys.",
    volume = "59",
    pages = "465",
    year = "1987"
}

@article{Churchill:2020uvk,
    author = "Churchill, Jessica and Yan, Li and Jeon, Sangyong and Gale, Charles",
    title = "{Emission of electromagnetic radiation from the early stages of relativistic heavy-ion collisions}",
    eprint = "2008.02902",
    archivePrefix = "arXiv",
    primaryClass = "hep-ph",
    doi = "10.1103/PhysRevC.103.024904",
    journal = "Phys. Rev. C",
    volume = "103",
    number = "2",
    pages = "024904",
    year = "2021"
}

@article{Garcia-Montero:2023lrd,
    author = {Garcia-Montero, Oscar and Mazeliauskas, Aleksas and Plaschke, Philip and Schlichting, S{\"o}ren},
    title = "{Pre-equilibrium photons from the early stages of heavy-ion collisions}",
    eprint = "2308.09747",
    archivePrefix = "arXiv",
    primaryClass = "hep-ph",
    doi = "10.1007/JHEP03(2024)053",
    journal = "JHEP",
    volume = "03",
    pages = "053",
    year = "2024"
}

@article{Garcia-Montero:2024lbl,
    author = {Garcia-Montero, Oscar and Plaschke, Philip and Schlichting, S{\"o}ren},
    title = "{Scaling of pre-equilibrium dilepton production in QCD kinetic theory}",
    eprint = "2403.04846",
    archivePrefix = "arXiv",
    primaryClass = "hep-ph",
    doi = "10.1103/PhysRevD.111.034036",
    journal = "Phys. Rev. D",
    volume = "111",
    number = "3",
    pages = "034036",
    year = "2025"
}

@article{Paquet:2015lta,
    author = {Paquet, Jean-Fran{\c{c}}ois and Shen, Chun and Denicol, Gabriel S. and Luzum, Matthew and Schenke, Bj{\"o}rn and Jeon, Sangyong and Gale, Charles},
    title = "{Production of photons in relativistic heavy-ion collisions}",
    eprint = "1509.06738",
    archivePrefix = "arXiv",
    primaryClass = "hep-ph",
    doi = "10.1103/PhysRevC.93.044906",
    journal = "Phys. Rev. C",
    volume = "93",
    number = "4",
    pages = "044906",
    year = "2016"
}

@article{Turbide:2003si,
    author = "Turbide, Simon and Rapp, Ralf and Gale, Charles",
    title = "{Hadronic production of thermal photons}",
    eprint = "hep-ph/0308085",
    archivePrefix = "arXiv",
    reportNumber = "NORDITA-2003-51-NP",
    doi = "10.1103/PhysRevC.69.014903",
    journal = "Phys. Rev. C",
    volume = "69",
    pages = "014903",
    year = "2004"
}

@article{Fries:2002kt,
    author = "Fries, Rainer J. and Muller, Berndt and Srivastava, Dinesh K.",
    title = "{High-energy photons from passage of jets through quark gluon plasma}",
    eprint = "nucl-th/0208001",
    archivePrefix = "arXiv",
    reportNumber = "DUKE-TH-02-224",
    doi = "10.1103/PhysRevLett.90.132301",
    journal = "Phys. Rev. Lett.",
    volume = "90",
    pages = "132301",
    year = "2003"
}

@article{Turbide:2005fk,
    author = "Turbide, Simon and Gale, Charles and Jeon, Sangyong and Moore, Guy D.",
    title = "{Energy loss of leading hadrons and direct photon production in evolving quark-gluon plasma}",
    eprint = "hep-ph/0502248",
    archivePrefix = "arXiv",
    doi = "10.1103/PhysRevC.72.014906",
    journal = "Phys. Rev. C",
    volume = "72",
    pages = "014906",
    year = "2005"
}

@article{Qin:2009bk,
    author = "Qin, Guang-You and Ruppert, Jorg and Gale, Charles and Jeon, Sangyong and Moore, Guy D.",
    title = "{Jet energy loss, photon production, and photon-hadron correlations at RHIC}",
    eprint = "0906.3280",
    archivePrefix = "arXiv",
    primaryClass = "hep-ph",
    doi = "10.1103/PhysRevC.80.054909",
    journal = "Phys. Rev. C",
    volume = "80",
    pages = "054909",
    year = "2009"
}

@article{Wu:2024vyc,
    author = "Wu, Xiang-Yu and Gao, Han and Forster, Bailey and Gale, Charles and Jackson, Greg and Jeon, Sangyong",
    title = "{Thermal Dilepton Polarization and Dynamics of the QCD Plasma in Relativistic Heavy-Ion Collisions}",
    eprint = "2412.15052",
    archivePrefix = "arXiv",
    primaryClass = "nucl-th",
    doi = "10.1103/c4jv-jxkn",
    journal = "Phys. Rev. Lett.",
    volume = "134",
    number = "24",
    pages = "242301",
    year = "2025"
}

@misc{Gao:2026vxs,
    author = "Gao, Han and Wu, Xiang-Yu and Gale, Charles and Jackson, Greg and Jeon, Sangyong",
    title = "{The polarization of thermal dileptons emitted in high-energy heavy-ion collisions}",
    eprint = "2605.30180",
    archivePrefix = "arXiv",
    primaryClass = "nucl-th",
    month = "5",
    year = "2026"
}

@article{Du:2025dot,
    author = "Du, Lipei and Heinz, Ulrich",
    title = "{Electromagnetic Tomography of Radial Flow in the Quark-Gluon Plasma}",
    eprint = "2505.20752",
    archivePrefix = "arXiv",
    primaryClass = "hep-ph",
    doi = "10.1103/ptyq-cs9m",
    journal = "Phys. Rev. Lett.",
    volume = "136",
    number = "10",
    pages = "102301",
    year = "2026"
}

@article{Churchill:2023zkk,
    author = "Churchill, Jessica and Du, Lipei and Gale, Charles and Jackson, Greg and Jeon, Sangyong",
    title = "{Virtual Photons Shed Light on the Early Temperature of Dense QCD Matter}",
    eprint = "2311.06951",
    archivePrefix = "arXiv",
    primaryClass = "nucl-th",
    doi = "10.1103/PhysRevLett.132.172301",
    journal = "Phys. Rev. Lett.",
    volume = "132",
    number = "17",
    pages = "172301",
    year = "2024"
}

@article{PHENIX:2014nkk,
    author = "Adare, A. and others",
    collaboration = "PHENIX",
    title = "{Centrality dependence of low-momentum direct-photon production in Au$+$Au collisions at $\sqrt{s_{_{NN}}}=200$ GeV}",
    eprint = "1405.3940",
    archivePrefix = "arXiv",
    primaryClass = "nucl-ex",
    doi = "10.1103/PhysRevC.91.064904",
    journal = "Phys. Rev. C",
    volume = "91",
    number = "6",
    pages = "064904",
    year = "2015"
}

@article{PHENIX:2015igl,
    author = "Adare, A. and others",
    collaboration = "PHENIX",
    title = "{Azimuthally anisotropic emission of low-momentum direct photons in Au$+$Au collisions at $\sqrt{s_{_{NN}}}=200$ GeV}",
    eprint = "1509.07758",
    archivePrefix = "arXiv",
    primaryClass = "nucl-ex",
    doi = "10.1103/PhysRevC.94.064901",
    journal = "Phys. Rev. C",
    volume = "94",
    number = "6",
    pages = "064901",
    year = "2016"
}

@article{STAR:2016use,
    author = "Adamczyk, L. and others",
    collaboration = "STAR",
    title = "{Direct virtual photon production in Au+Au collisions at $\sqrt{s_{NN}}$ = 200 GeV}",
    eprint = "1607.01447",
    archivePrefix = "arXiv",
    primaryClass = "nucl-ex",
    doi = "10.1016/j.physletb.2017.04.050",
    journal = "Phys. Lett. B",
    volume = "770",
    pages = "451--458",
    year = "2017"
}

@article{PHENIX:2022rsx,
    author = "Abdulameer, N. J. and others",
    collaboration = "PHENIX",
    title = "{Nonprompt direct-photon production in Au+Au collisions at sNN=200 GeV}",
    eprint = "2203.17187",
    archivePrefix = "arXiv",
    primaryClass = "nucl-ex",
    doi = "10.1103/PhysRevC.109.044912",
    journal = "Phys. Rev. C",
    volume = "109",
    number = "4",
    pages = "044912",
    year = "2024"
}

@article{ALICE:2015xmh,
    author = "Adam, Jaroslav and others",
    collaboration = "ALICE",
    title = "{Direct photon production in Pb-Pb collisions at $\sqrt{s_{NN}} =$ 2.76 TeV}",
    eprint = "1509.07324",
    archivePrefix = "arXiv",
    primaryClass = "nucl-ex",
    reportNumber = "ALICE-PUBLIC-2015-007, CERN-PH-EP-2015-254",
    doi = "10.1016/j.physletb.2016.01.020",
    journal = "Phys. Lett. B",
    volume = "754",
    pages = "235--248",
    year = "2016"
}

@article{Gale:2021emg,
    author = {Gale, Charles and Paquet, Jean-Fran{\c{c}}ois and Schenke, Bj{\"o}rn and Shen, Chun},
    title = "{Multimessenger heavy-ion collision physics}",
    eprint = "2106.11216",
    archivePrefix = "arXiv",
    primaryClass = "nucl-th",
    doi = "10.1103/PhysRevC.105.014909",
    journal = "Phys. Rev. C",
    volume = "105",
    number = "1",
    pages = "014909",
    year = "2022"
}

@article{Arnold:2001ms,
    author = "Arnold, Peter Brockway and Moore, Guy D. and Yaffe, Laurence G.",
    title = "{Photon emission from quark gluon plasma: Complete leading order results}",
    eprint = "hep-ph/0111107",
    archivePrefix = "arXiv",
    reportNumber = "UW-PT-01-22",
    doi = "10.1088/1126-6708/2001/12/009",
    journal = "JHEP",
    volume = "12",
    pages = "009",
    year = "2001"
}

@article{Lu:2025yry,
    author = "Lu, Xingjian and Shi, Shuzhe",
    title = "{Spectral Bogoliubov-Born-Green-Kirkwood-Yvon hierarchical approach: A scalable scheme for nonlinear Boltzmann and correlation kinetics}",
    eprint = "2507.14243",
    archivePrefix = "arXiv",
    primaryClass = "nucl-th",
    doi = "10.1103/zxkx-5t8k",
    journal = "Phys. Rev. C",
    volume = "113",
    number = "3",
    pages = "034917",
    year = "2026"
}

@misc{Lu:2026nmp,
    author = "Lu, Xingjian and Shi, Shuzhe",
    title = "{Probing Pair Correlations in QCD Matter with Photon Spectra}",
    eprint = "2606.02414",
    archivePrefix = "arXiv",
    primaryClass = "hep-ph",
    month = "6",
    year = "2026"
}

@article{Shen:2014vra,
    author = "Shen, Chun and Qiu, Zhi and Song, Huichao and Bernhard, Jonah and Bass, Steffen and Heinz, Ulrich",
    title = "{The iEBE-VISHNU code package for relativistic heavy-ion collisions}",
    eprint = "1409.8164",
    archivePrefix = "arXiv",
    primaryClass = "nucl-th",
    doi = "10.1016/j.cpc.2015.08.039",
    journal = "Comput. Phys. Commun.",
    volume = "199",
    pages = "61--85",
    year = "2016"
}

@article{Shi:2018izg,
    author = "Shi, Shuzhe and Liao, Jinfeng and Gyulassy, Miklos",
    title = "{Global constraints from RHIC and LHC on transport properties of QCD fluids in CUJET/CIBJET framework}",
    eprint = "1808.05461",
    archivePrefix = "arXiv",
    primaryClass = "hep-ph",
    doi = "10.1088/1674-1137/43/4/044101",
    journal = "Chin. Phys. C",
    volume = "43",
    number = "4",
    pages = "044101",
    year = "2019"
}

\appendix

\section{Four-column inverse fit}
\label{app:inverse}

We select one set of four columns for all collision classes by forward selection.
The candidate pool has $73$ connected modes, each multiplied by either $q_{\rm C}$ or $q_T$, for $146$ columns.
We perform the selection with both conditions A and B imposed as hard constraints.
At each step, we screen the remaining candidates on a coarse constraint grid, refit the six best candidates on the full constraint grid, and add the one with the smallest sum of the four class objectives.
The corrected Akaike criterion is smallest at two columns on this selection path, but we retain four so that one direction remains free after the three active hard constraints are imposed.
The selected columns are the modes $(0,0,3)$ and $(1,0,2)$ with both $q_{\rm C}$ and $q_T$.
We use this basis in every constraint setting to compare the fits directly.
Within each setting, we fit the coefficients separately for each class by minimizing Eq.~\eqref{eq:chi2_joint_v2} with the corresponding positivity constraints or penalties and a small ridge regularizer.

The two lowest ALICE $20-40\%$ yield points are $95\%$ confidence-level upper limits, so they are omitted from the ordinary yield-residual sum in Eq.~\eqref{eq:chi2_joint_v2}.
They add the one-sided term
\begin{equation*}
 \chi^2_{\rm UL}(\boldsymbol\psi)
 =\sum_{i\in\mathrm{UL}}
 \left[
  \frac{\max\{0,Y_i(\boldsymbol\psi)-Y_i^{\rm UL}\}}{Y_i^{\rm UL}/1.645}
 \right]^2 .
\end{equation*}
Here $Y_i^{\rm UL}$ is the reported upper limit, and $Y_i^{\rm UL}/1.645$ is the adopted one-sided Gaussian width.
The term vanishes when the predicted yield is below the limit and penalizes only predictions above it.

\end{document}